\documentclass[runningheads]{llncs}
\usepackage{graphicx}
\usepackage{amsmath,amssymb,amsfonts}

\begin{document}
\title{Are Direct Links Necessary in Random Vector Functional Link Networks for Regression?\thanks{Supported by Grant 2017/27/B/ST6/01804 from the National Science Centre, Poland.}}
\titlerunning{Are Direct Links Necessary in RVFL NNs for Regression?}
	%
\author{Grzegorz Dudek}

	%
\authorrunning{G. Dudek}
%
\institute{Electrical Engineering Faculty, Częstochowa University of Technology,\\ Częstochowa, Poland\\
\email{dudek@el.pcz.czest.pl}}
%
\maketitle              
\begin{abstract}

A random vector functional link network (RVFL) is widely used as a universal approximator for classification and regression problems. The big advantage of RVFL is fast training without backpropagation. This is because the weights and biases of hidden nodes are selected randomly and stay untrained. Recently, alternative architectures with randomized learning are developed which differ from RVFL in that they have no direct links and a bias term in the output layer. In this study, we investigate the effect of direct links and output node bias on the regression performance of RVFL. For generating random parameters of hidden nodes we use the classical method and two new methods recently proposed in the literature. We test the RVFL performance on several function approximation problems with target functions of different nature: nonlinear, nonlinear with strong fluctuations, nonlinear with linear component and linear. Surprisingly, we found that the direct links and output node bias do not play an important role in improving RVFL accuracy for typical nonlinear regression problems.

\keywords{Random vector functional link network \and Neural networks with random hidden nodes \and Randomized learning algorithms.}
\end{abstract}
\section{Introduction}

A random vector functional link network (RVFL) is a type of feedforward neural network (FNN) with a single hidden layer and direct links between input and output layers. Unlike typical FNN, in RVFL the weights and biases of the hidden nodes are selected randomly and stay fixed. The only parameters which are learned are weights and biases of the output layer. Due to randomization in RVFL we can avoid complicated and time-consuming gradient descent methods for solving the optimization problem which is non-convex in typical FNNs. It is commonly known that the gradient learning methods have many drawbacks such as sensitivity to initial values of parameters, convergence to local minima, vanishing/exploding gradients in deep neural structures, and usually additional hyperparameters to tune. In RVFL the resulting optimization problem becomes convex and the output weights can be determined analytically by using a simple standard linear least-squares method \cite{Pri15}.

RVFL is extensively used for classification and regression problems due to its adaptive nature and universal approximation property. Many simulation studies reported in the literature show the high performance of the randomized models which is compared
to fully adaptable ones. Randomization which is cheaper than optimization ensures faster training and simpler implementation. 

RVFL is not the only FNN solution with randomization. Alternative approaches such as \cite{Sch92} and many other new solutions do not have direct links between the input and output layers \cite{Wan17}. The effect of direct links as well as a bias in the output layer on the RVFL performance in classification tasks was investigated in \cite{Zha16}. The basic conclusion of that work was that the direct link plays an important performance enhancing role in RVFL, while the bias term in the output neuron had no significant effect. In this work, we investigate the effect of direct links and output node bias on the regression performance of RVFL. For generating random parameters of hidden nodes we use the classical method and two new methods recently proposed in the literature. We test the RVFL performance on several function approximation problems with target functions of different nature: nonlinear, nonlinear with strong fluctuations, nonlinear with linear component and linear. 

The remainder of this paper is structured as follows. In Section 2, we briefly present RVFL learning algorithm and the decomposition of the function built by RVFL. In Section 3, we describe three methods of generating weights and biases of hidden nodes. Section 4 reports the simulation study and compares results for different RVFL configurations, different methods of random parameters generation, and different regression problems. Finally, Section 5 concludes the work.

\section{Random Vector Functional Link Network}

\begin{figure}[htbp]
	\centering
	\includegraphics[width=0.4\textwidth]{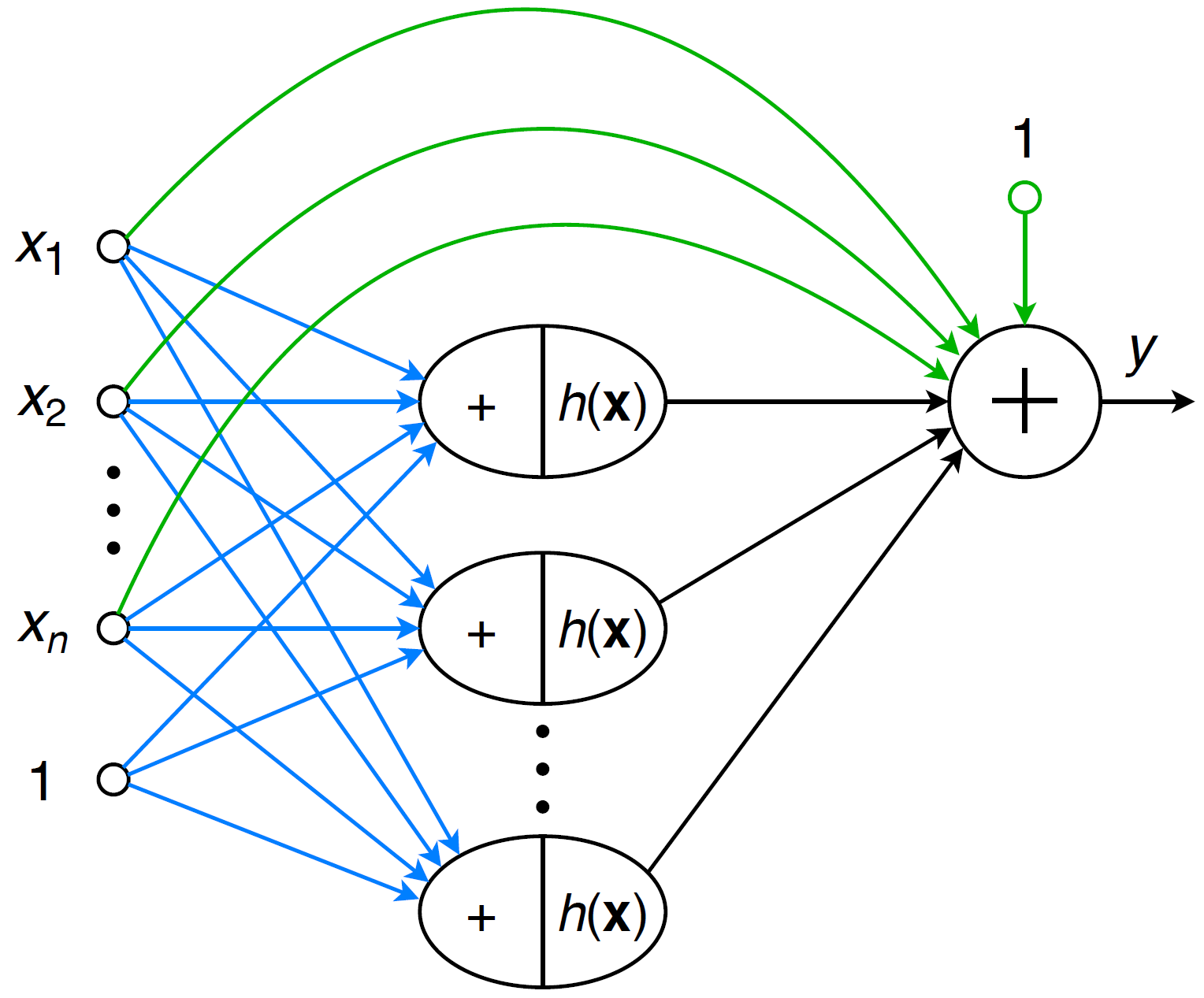}
	\caption{RVFL architecture (random links in blue, direct links and output node bias in green).}
	\label{figA}
\end{figure}

RVFL was proposed by Pao and Takefuji \cite{Pao92}. It was proven in \cite{Ige95} that RVFL is a universal approximator for a continuous function on a bounded finite dimensional set with a closed-form solution. RVFL can be regarded as a single hidden layer FNN built with a specific randomized algorithm. The RVFL architecture is shown in Fig. \ref{figA}. Note that in addition to a hidden layer transforming inputs nonlinearly, RVFL also has direct links connecting an input layer with output nodes. The weights and biases of hidden nodes, $a_{i,j}, b_i$, respectively, are randomly assigned and fixed during the training phase. The output weights, $\beta_i$, are analytically evaluated using a linear least-square method. This results in a flat-net architecture for which only weights $\beta_i$ must be learned. The learning problem, which is non-convex for the full learning of all parameters, becomes convex in RVFL. So, the time-consuming gradient-based learning algorithms are not needed, which makes the learning process much easier to implement and extremely rapid.

The learning algorithm of RVFL ia as follows. One output is considered, $ m $ hidden nodes and $ n $ inputs. The training set is $ \Phi = \{ (\mathbf{x}_l, y_l) | \mathbf{x}_l \in \mathbb{R}^n, y_l \in \mathbb{R}, l = 1, 2, ..., N \} $ and the activation function of hidden nodes is $ h(\mathbf{x}) $, which is nonlinear piecewise continuous function, e.g. a~sigmoid:
\begin{equation}
h(\mathbf{x}) = \frac{1}{1 + \exp\left(-\left(\mathbf{a}^T\mathbf{x}+b\right)\right)}
\end{equation}
 
\begin{enumerate}
	\item 
	Randomly generate hidden node parameters: weights $ \mathbf{a}_i = \left[ a_{i,1}, a_{i,2}, \ldots, a_{i,n}\right]^T $ and biases $ b_i$ for all nodes, $i = 1, 2, ..., m $, according to any continuous sampling distribution.
	
	\item 
	Calculate the hidden layer output matrix $ \mathbf{H} $:
	\begin{equation}\label{key}
	\mathbf{H} = \left[
	\begin{array}{c}
	\mathbf{h}(\mathbf{x}_1) \\
	\vdots \\
	\mathbf{h}(\mathbf{x}_N) 
	\end{array}
	\right] =
	\left[
	\begin{array}{ccc}
	h_1(\mathbf{x}_1) & \ldots & h_m(\mathbf{x}_1) \\
	\vdots & \vdots & \vdots \\
	h_1(\mathbf{x}_N) & \ldots & h_m(\mathbf{x}_N)
	\end{array}
	\right]
	\end{equation}
	where $ h_i(\mathbf{x}) $ is an activation function of the $ i $-th node. 
	
	The $ i $-th column of $ \mathbf{H} $ is the $ i $-th hidden node output vector with respect to inputs $ \mathbf{x}_1, \mathbf{x}_2, \ldots, \mathbf{x}_N $. Hidden nodes map nonlinearly inputs from $n$-dimensional input space to $ m $-dimensional space. The output matrix $ \mathbf{H} $ remains unchanged because parameters of hidden nodes, $ \mathbf{a}_i $ and $ b_i $, are fixed.
	
	\item 
	Calculate the output weights:
	\begin{equation}
	\boldsymbol{\beta} = [\mathbf{1} \, \mathbf{X} \, \mathbf{H}]^+\mathbf{Y}
	\end{equation}
	where $ \boldsymbol{\beta} = [\beta_0, \beta_1, ..., \beta_{n+m}]^T $ is a~vector of output weights,	$\mathbf{1}$ is an $N\times1$ one vector corresponding to an output node bias, $\mathbf{X}$ is a $N\times n $ input matrix, $\mathbf{Y} = [y_1, y_2, ..., y_N]^T $ is a~vector of target outputs, and $ [.]^+ $ is the Moore-Penrose generalized inverse of matrix $ [.] $.
			
\end{enumerate}

The above equation for $ \boldsymbol{\beta} $ results from the following criterion for minimizing the approximation error:
\begin{equation}
\min\left\| [\mathbf{1} \, \mathbf{X} \, \mathbf{H}]\boldsymbol{\beta}-\mathbf{Y} \right\|
\end{equation}
	
A function expressed by RVFL is a~linear combination of inputs $x_i$ and activation functions $ h_i(\mathbf{x}) $:

\begin{equation}\label{eqf}
f(\mathbf{x}) = \sum_{i=1}^{n}\beta_i x_i + \sum_{i=1}^{m}\beta_{n+i}h_i(\mathbf{x}) + \beta_0
\end{equation}

\begin{figure}[htbp]
	\centering
	\includegraphics[width=0.32\textwidth]{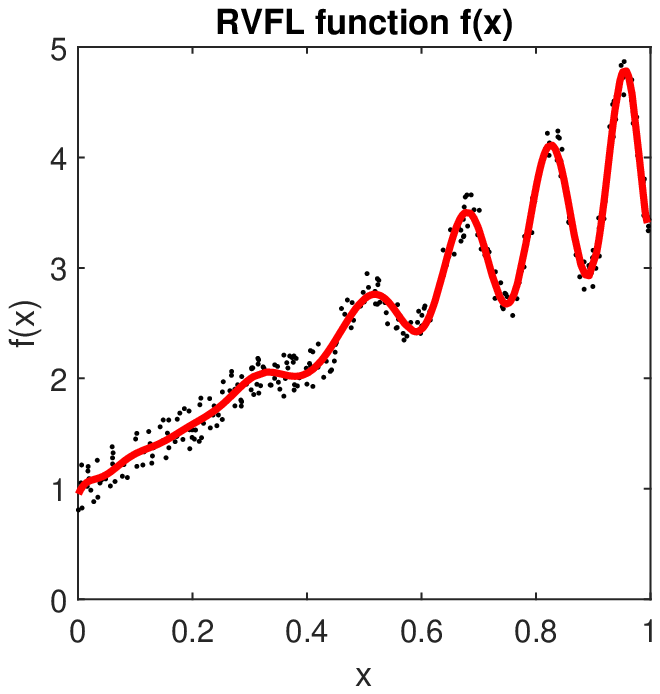}
	\includegraphics[width=0.32\textwidth]{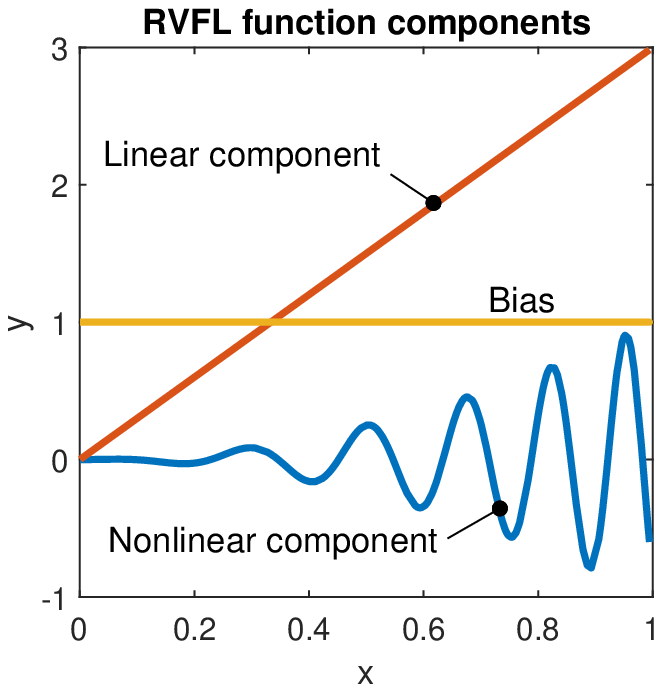}
	\includegraphics[width=0.32\textwidth]{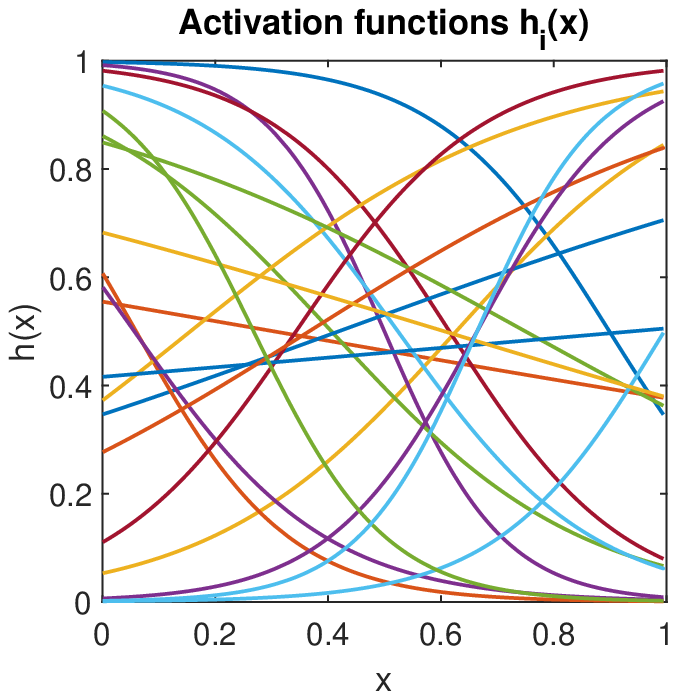}
	\caption{A example of RVFL function $f(\mathbf{x})$ (left panel), its decomposition (middle panel) and sigmoids of hidden nodes constructing a nonlinear component (right panel).}
	\label{figC}
\end{figure}

Note that the first component in \eqref{eqf} is linear and represents a hyperplane, the second component expresses a nonlinear function and the last component is a bias. These three components of function $f(\mathbf{x})$ are depicted in Fig. \ref{figC}. The nonlinear component is a linear combination of hidden node activation functions $h_i(\mathbf{x})$ (sigmoids in our case) which are also shown in this figure. 

A natural question that arises is: are all these three components of function $f(\mathbf{x})$ necessary for an approximation of the target function? Is only a nonlinear component not enough? In the experimental part of this work, we try to answer these questions.  

\section{Generating Weights and Biases of Hidden Nodes}

The key issue in FNN randomized learning is finding a way of generating the random hidden node parameters to obtain a good projection space \cite{Dud19b}. The standard approach is to generate both weights and biases randomly with a fixed interval from any continuous
sampling distribution. The symmetric interval ensures a universal approximation property for the functions which meet Lipschitz condition \cite{Hus99}. The appropriate selection of this interval is a problem that has not been solved as yet and is considered to be one of the major research challenges in the area of FNN randomized learning \cite{Zha16b}, \cite{Cao18}. In many cases the interval is selected as $[-1, 1]$ without any justification, regardless of the problem solved, data distribution, and activation functions type. In practical applications, the optimization of this interval is recommended for better model performance \cite{Pao94}, \cite{Hus99}, \cite{Wan17}. 

In the experimental part of this work, we use three methods of generating random parameters. One of them is a standard approach, where both weights and biases of hidden nodes are generated uniformly from interval $[-u,u]$. A bound of this symmetrical interval, $u$, is adjusted to the target function (TF). This method of generating random parameters is denoted as Gs. Note that in the right panel of Fig. \ref{figC} the sigmoids are randomly evenly distributed in the input interval which is a correct solution. Unfortunately, the Gs method does not ensure such even distribution (see \cite{Dud19b}).          

Another method (denoted as Gu in this work) was proposed in \cite{Dud19}. Due to different functions of the hidden node parameters, i.e. weights express slopes of the sigmoids and biases express their shifts, they should be generated separately, not both from the same interval. According to Gu method, first, the weights $a_{i,j}$ are selected randomly from $U(-u,u)$ and then biases are determined as follows:      

\begin{equation}
b_i = -\mathbf{a}^T_i\mathbf{x}^*_i
\label{eqb}
\end{equation}
where $\mathbf{x}^*_i$ is one of the training point selected randomly (see \cite{Dud19} for other variants). 

Determining biases from \eqref{eqb} ensures that the hidden nodes will be placed in accordance with the input data density \cite{Tyu06}. The Gu method ensures that all sigmoids have their steepest fragments, which are most useful for modeling TF fluctuations, inside the input hypercube as shown in the right panel of Fig. \ref{figC}. In this way, Gu improves a drawback of Gs which can generate sigmoids having their saturated fragments inside the input hypercube. These fragments are useless for building a nonlinear fitted function. Moreover, in Gs, it is difficult to adjust both parameters, weights and biases, when they are selected from the same interval. Gu selects weights first and then calculates biases depending on the weights and data distribution.      

The third method of generating random parameters of hidden nodes ensures sigmoids with uniformly distributed slope angles \cite{Dud19c}, \cite{Dud19}. This method is denoted as G$\alpha$ in this work. In many cases G$\alpha$ gives better performance of the model than Gu, especially for highly nonlinear TFs (see \cite{Dud19} for comparison of Gs, Gu and G$\alpha$). In the first step, G$\alpha$ generates slope angles of sigmoids $|\alpha_{i,j}| \sim U(\alpha_{\min}, \alpha_{\max})$, where $\alpha_{\min} \in (0^\circ, 90^\circ)$ and $\alpha_{\max} \in (\alpha_{\min}, 90^\circ)$. The bound angles, $\alpha_{\min}$ and $\alpha_{\max}$, are tuned to the TF. For highly nonlinear TFs, with strong fluctuations, only $\alpha_{\min}$ can be adjusted, keeping $\alpha_{\max}=90^\circ$. The weights are calculated on the basis of the angles from:

\begin{equation}
a_{i,j}=4 \tan \alpha_{i,j} 
\label{eq6a}
\end{equation}

G$\alpha$ ensures random slopes between $\alpha_{\min}$ and $\alpha_{\max}$ for the multidimensional sigmoids in each of $n$ directions. The biases of the hidden nodes are calculated from \eqref{eqb} to set the sigmoids inside the input hypercube depending on data density.  

\section{Experiments and Results}

In this section, to asses the impact of the direct links and bias in the output node on RVFL performance we consider the following RVFL configurations:
\begin{description}
	\item [+dl+b] -- RVFL with direct links and output node bias,
	
	\item [+dl--b] -- RVFL with direct links and without output node bias,

	\item [--dl+b] -- RVFL without direct links and with output node bias,

	\item [--dl--b] -- RVFL without direct links and output node bias.
	
\end{description}

We use sigmoids as activation functions. The hidden node weights and biases are generated using three methods described in Section 3:   

\begin{description}
	\item [Gs] -- the standard approach of generating both weights and biases from $U(-u,u)$,
	\item [Gu] -- generating weights from $U(-u,u)$ and biases according to \eqref{eqb},
	\item [G$\alpha$] -- generating slope angles of sigmoids $|\alpha_{i,j}| \sim U(\alpha_{\min}, \alpha_{\max})$, then calculating weights from \eqref{eq6a}, and biases from \eqref{eqb}.  
\end{description} 

The parameters of these methods as well as the number of hidden nodes $m$ were selected in grid search for each RVFL variant and TF from the sets: $m = \{1, 2, ..., 10, 20, ..., 100, 200, ..., 1000\}$, $u = \{1, 2, ..., 10, 20, 50,100\}$ for 2-dimensional data or $u = \{0.1, 0.2, ...,1, 2, ..., 5\}$ for 5 and 10-dimensional data, $\alpha_{\min} = \{0, 15, ..., 75\}$, and $\alpha_{\max} = \{\alpha_{\min}+15, \alpha_{\min}+30,..., 90\}$.

We test RVFL performance over several regression problems using TFs defined as:

\begin{equation}
g(\mathbf{x}) = \exp\left(-\sum_{j=1}^{n}(x_j - 0.5)^2\right)
\label{eqTF1}
\end{equation}

\begin{equation}
g(\mathbf{x}) = \alpha\sum_{j=1}^{n}\sin\left(20\cdot\exp x_j\right)\cdot x_j^2 + \delta\cdot 3\sum_{j=1}^{n}x_j
\label{eqTF2}
\end{equation}
where $\alpha$ and $\delta$ are $0/1$ variables.

\begin{figure}[htbp]
	\centering
	\includegraphics[width=0.32\textwidth]{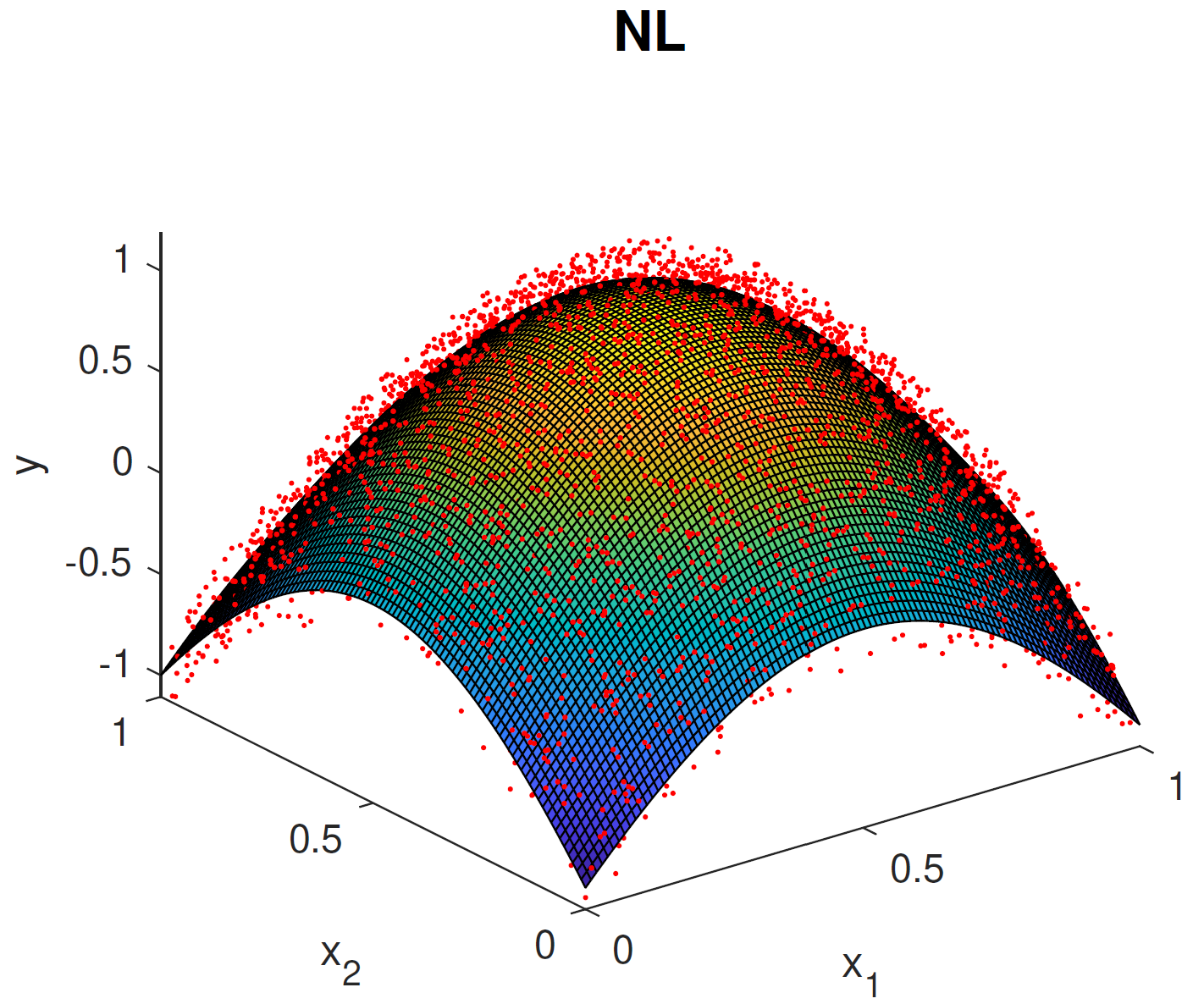}
	\includegraphics[width=0.32\textwidth]{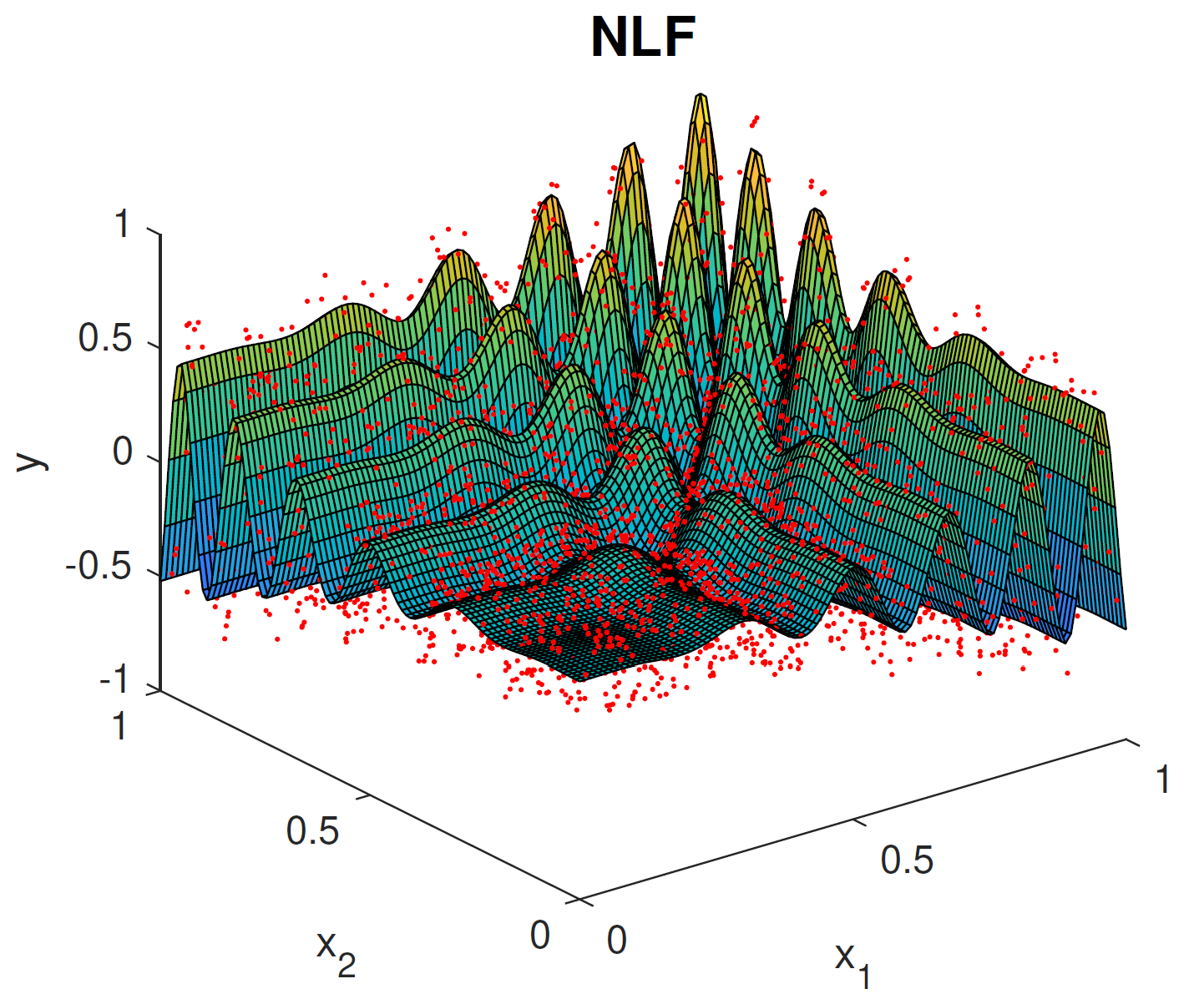}
	\includegraphics[width=0.32\textwidth]{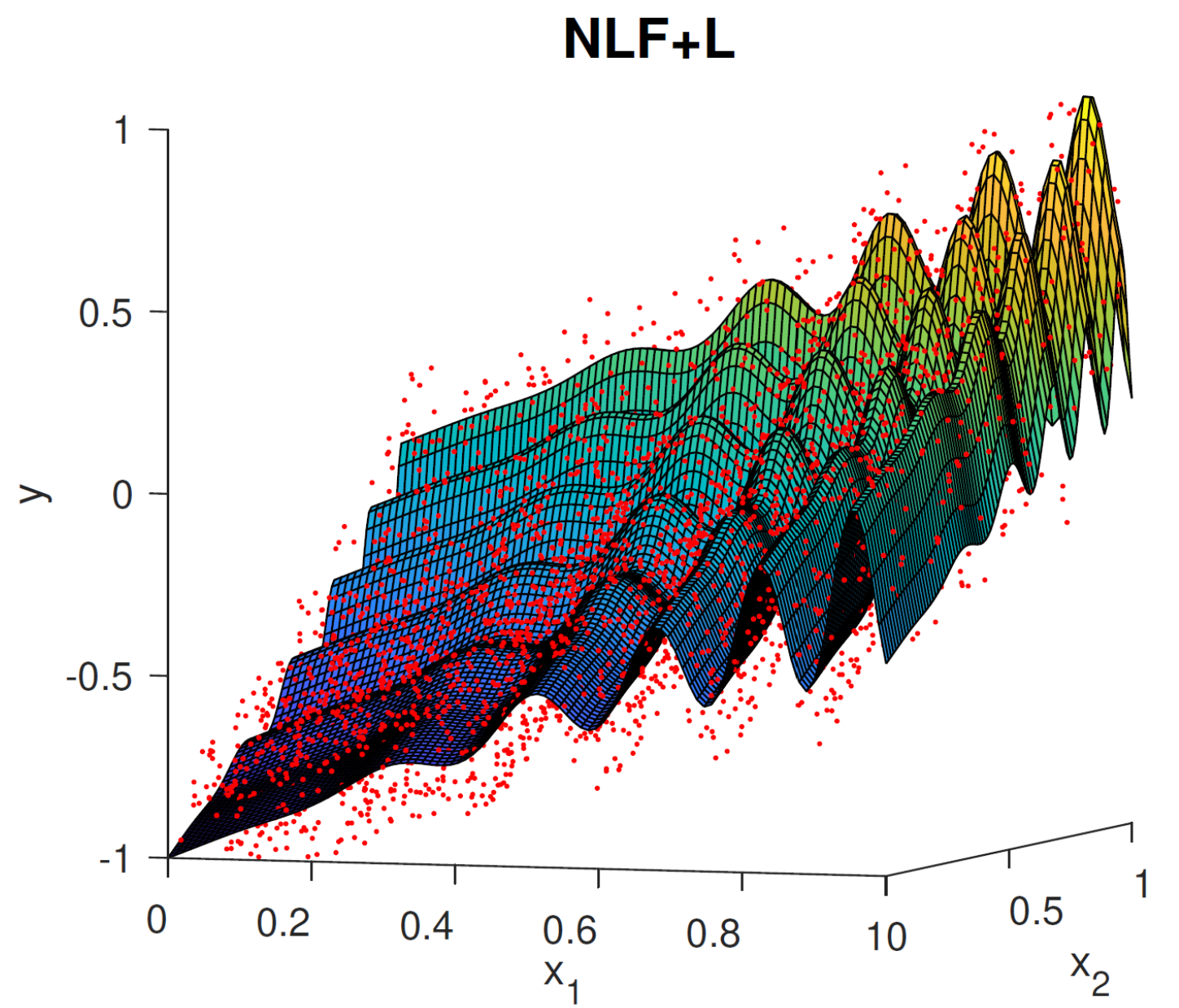}
	\caption{Target functions and training points for $n=2$.}
	\label{figF}
\end{figure}

Function \eqref{eqTF1} is a simple nonlinear function shown in the left panel of Fig. \ref{figF}. The first component of function \eqref{eqTF2} is a highly nonlinear function shown in the middle panel of Fig. \ref{figF}. The second component is a hyperplane. The TF can be composed of these both components if $\alpha=1$ and $\delta=1$ or of only one component if $\alpha=0$ or $\delta=0$. The TF with both components is shown in the right panel of Fig. \ref{figF}. To asses the RVFL regression performance on TFs of different character four types of TFs were used:

\begin{description}
	\item [NL] -- nonlinear \eqref{eqTF1},
	
	\item [NLF] -- nonlinear with strong fluctuations, \eqref{eqTF2} with $\alpha=1$ and $\delta=0$,
	
	\item [NLF+L] -- nonlinear with fluctuations and a linear component, \eqref{eqTF2} with $\alpha=1$ and $\delta=1$,
	
	\item [L] -- linear, \eqref{eqTF2} with $\alpha=0$ and $\delta=1$.

\end{description}

The experiments were carried out for $n=2$ and $N=5000$, $n=5$ and $N=20000$, and $n=10$ and $N=50000$. As an accuracy measure, we used root mean squares error (RMSE). In each case, RFVL networks were trained 100 times and the final errors were calculated as the averages over 100 trials.

Tables 1--4 show RMSE for different TFs and RVFL variants. To confirm the significance of error differences between RVFL without direct links and output node bias (configuration --dl--b) and other RVFL configurations we used a two-sided Wilcoxon signed-rank test. We performed the tests separately for Gs, Gu and G$\alpha$. The null hypothesis was as follows: $d = RMSE_{-dl-b}-RMSE_{v}$, where $v$ is $+dl+b, +dl-b$ or $-dl+b$, respectively, comes from a distribution with zero median. It was assumed that $p$-value below 5\% indicates a rejection of the null hypothesis. The cases of the null hypothesis rejection are underlined in the tables (i.e. the cases +dl+b, +dl--b or --dl+b for which the error was significantly lower than for --dl--b). 

\begin{table}[]
	\setlength {\tabcolsep}{5pt}
	\caption{RMSE for NL.}
	\begin{center}
		\begin{tabular}{|cc|c|c|c|}
			\hline
			\multicolumn{2}{|c|}{RVFL variant} & $n = 2$     & $n = 5$  & $n = 10$   \\ \hline
			
			Gs	&+dl+b&	7.50E-03	$\pm$	9.39E-04	&	0.0121	$\pm$	5.29E-04	&	0.0236	$\pm$	4.72E-04	\\	
			&+dl--b&	7.41E-03	$\pm$	9.21E-04	&	0.0121	$\pm$	5.33E-04	&	0.0236	$\pm$	4.70E-04	\\	
			&--dl+b&	7.30E-03	$\pm$	9.42E-04	&	0.0121	$\pm$	5.28E-04	&	0.0237	$\pm$	4.46E-04	\\	
			&--dl--b&	7.20E-03	$\pm$	9.11E-04	&	0.0121	$\pm$	5.22E-04	&	0.0237	$\pm$	4.44E-04	\\	\hline
			Gu	&+dl+b&	7.45E-03	$\pm$	9.07E-04	&	0.0128	$\pm$	5.05E-04	&	\underline{0.0227}	$\pm$	4.10E-04	\\	
			&+dl--b&	7.38E-03	$\pm$	9.14E-04	&	0.0128	$\pm$	5.13E-04	&	0.0229	$\pm$	4.21E-04	\\	
			&--dl+b&	7.28E-03	$\pm$	9.75E-04	&	0.0128	$\pm$	5.16E-04	&	\underline{0.0227}	$\pm$	4.07E-04	\\	
			&--dl--b&	7.20E-03	$\pm$	9.95E-04	&	0.0128	$\pm$	5.15E-04	&	0.0230	$\pm$	4.23E-04	\\	\hline
			G$\alpha$	&+dl+b&	7.47E-03	$\pm$	9.39E-04	&	0.0130	$\pm$	5.05E-04	&	0.0217	$\pm$	4.11E-04	\\	
			&+dl--b&	7.39E-03	$\pm$	9.31E-04	&	0.0130	$\pm$	5.10E-04	&	0.0219	$\pm$	4.20E-04	\\	
			&--dl+b&	7.27E-03	$\pm$	9.53E-04	&	0.0129	$\pm$	4.92E-04	&	0.0217	$\pm$	4.06E-04	\\	
			&--dl--b&	7.16E-03	$\pm$	9.87E-04	&	0.0130	$\pm$	4.98E-04	&	0.0219	$\pm$	4.20E-04	\\	\hline

		\end{tabular}
		\label{tab1}
	\end{center}
\end{table}

\begin{table}[]
	\setlength {\tabcolsep}{5pt}
	\caption{RMSE for NLF.}
	\begin{center}
		\begin{tabular}{|cc|c|c|c|}
			\hline
			\multicolumn{2}{|c|}{RVFL variant} & $n = 2$     & $n = 5$  & $n = 10$   \\ \hline
			Gs	&+dl+b&	0.0414	$\pm$	0.0055	&	0.2268	$\pm$	0.0122	&	0.2203	$\pm$	0.0098	\\	
			&+dl--b&	0.0414	$\pm$	0.0055	&	0.2268	$\pm$	0.0122	&	0.2203	$\pm$	0.0098	\\	
			&--dl+b&	0.0415	$\pm$	0.0056	&	0.2268	$\pm$	0.0122	&	0.2203	$\pm$	0.0098	\\	
			&--dl--b&	0.0415	$\pm$	0.0056	&	0.2268	$\pm$	0.0122	&	0.2203	$\pm$	0.0098	\\	\hline
			Gu	&+dl+b&	0.0378	$\pm$	0.0028	&	0.2268	$\pm$	0.0121	&	0.2203	$\pm$	0.0098	\\	
			&+dl--b&	0.0378	$\pm$	0.0028	&	0.2268	$\pm$	0.0121	&	0.2203	$\pm$	0.0098	\\	
			&--dl+b&	0.0379	$\pm$	0.0027	&	0.2268	$\pm$	0.0121	&	0.2203	$\pm$	0.0098	\\	
			&--dl--b&	0.0379	$\pm$	0.0027	&	0.2268	$\pm$	0.0121	&	0.2203	$\pm$	0.0098	\\	\hline
			G$\alpha$	&+dl+b&	0.0335	$\pm$	0.0021	&	0.1702	$\pm$	0.0111	&	0.2026	$\pm$	0.0099	\\	
			&+dl--b&	0.0335	$\pm$	0.0021	&	0.1702	$\pm$	0.0111	&	0.2026	$\pm$	0.0099	\\	
			&--dl+b&	0.0336	$\pm$	0.0022	&	0.1704	$\pm$	0.0111	&	0.2030	$\pm$	0.0099	\\	
			&--dl--b&	0.0336	$\pm$	0.0022	&	0.1704	$\pm$	0.0111	&	0.2030	$\pm$	0.0099	\\	\hline

		\end{tabular}
		\label{tab2}
	\end{center}
\end{table}

\begin{table}[]
	\setlength {\tabcolsep}{5pt}
	\caption{RMSE for NLF+L.}
	\begin{center}
		\begin{tabular}{|cc|c|c|c|}
			\hline
			\multicolumn{2}{|c|}{RVFL variant} & $n = 2$     & $n = 5$  & $n = 10$   \\ \hline
			Gs	&+dl+b&	0.0375	$\pm$	0.0044	&	0.0887	$\pm$	0.0030	&	0.0802	$\pm$	0.0030	\\	
			&+dl--b&	0.0375	$\pm$	0.0044	&	0.0887	$\pm$	0.0030	&	0.0802	$\pm$	0.0030	\\	
			&--dl+b&	0.0374	$\pm$	0.0043	&	0.0888	$\pm$	0.0030	&	0.0803	$\pm$	0.0030	\\	
			&--dl--b&	0.0374	$\pm$	0.0043	&	0.0888	$\pm$	0.0030	&	0.0803	$\pm$	0.0030	\\	\hline
			Gu	&+dl+b&	0.0351	$\pm$	0.0019	&	0.0887	$\pm$	0.0030	&	0.0802	$\pm$	0.0030	\\	
			&+dl--b&	0.0351	$\pm$	0.0019	&	0.0887	$\pm$	0.0030	&	0.0802	$\pm$	0.0030	\\	
			&--dl+b&	0.0351	$\pm$	0.0019	&	0.0888	$\pm$	0.0030	&	0.0802	$\pm$	0.0030	\\	
			&--dl--b&	0.0351	$\pm$	0.0019	&	0.0888	$\pm$	0.0030	&	0.0802	$\pm$	0.0030	\\	\hline
			G$\alpha$	&+dl+b&	0.0307	$\pm$	0.0018	&	0.0706	$\pm$	0.0033	&	\underline{0.0750}	$\pm$	0.0028	\\	
			&+dl--b&	0.0307	$\pm$	0.0018	&	0.0706	$\pm$	0.0033	&	\underline{0.0754}	$\pm$	0.0028	\\	
			&--dl+b&	0.0308	$\pm$	0.0018	&	0.0707	$\pm$	0.0033	&	0.0762	$\pm$	0.0028	\\	
			&--dl--b&	0.0308	$\pm$	0.0018	&	0.0707	$\pm$	0.0033	&	0.0763	$\pm$	0.0028	\\	\hline

		\end{tabular}
		\label{tab2}
	\end{center}
\end{table}

\begin{table}[]
	\setlength {\tabcolsep}{5pt}
	\caption{RMSE for L.}
	\begin{center}
		\begin{tabular}{|cc|c|c|c|}
			\hline
			\multicolumn{2}{|c|}{RVFL variant} & $n = 2$     & $n = 5$  & $n = 10$   \\ \hline
			Gs	&+dl+b&	\underline{2.61E-03}	$\pm$	1.10E-03	&	1.89E-03	$\pm$	5.40E-04	&	\underline{1.52E-03}	$\pm$	3.80E-04	\\	
			&+dl--b&	\underline{2.62E-03}	$\pm$	1.09E-03	&	1.89E-03	$\pm$	5.40E-04	&	\underline{1.70E-03}	$\pm$	4.10E-04	\\	
			&--dl+b&	3.97E-03	$\pm$	9.87E-04	&	1.98E-03	$\pm$	5.20E-04	&	\underline{2.20E-03}	$\pm$	4.30E-04	\\	
			&--dl--b&	4.08E-03	$\pm$	9.63E-04	&	1.99E-03	$\pm$	5.20E-04	&	2.33E-03	$\pm$	3.46E-04	\\	\hline
			Gu	&+dl+b&	\underline{2.61E-03}	$\pm$	1.10E-03	&	1.89E-03	$\pm$	5.40E-04	&	\underline{1.52E-03}	$\pm$	3.80E-04	\\	
			&+dl--b&	\underline{2.72E-03}	$\pm$	1.05E-03	&	1.89E-03	$\pm$	5.40E-04	&	\underline{1.70E-03}	$\pm$	4.10E-04	\\	
			&--dl+b&	\underline{3.38E-03}	$\pm$	9.94E-04	&	1.93E-03	$\pm$	5.26E-04	&	\underline{2.02E-03}	$\pm$	4.45E-04	\\	
			&--dl--b&	3.78E-03	$\pm$	1.04E-03	&	1.93E-03	$\pm$	5.32E-04	&	2.26E-03	$\pm$	3.77E-04	\\	\hline
			G$\alpha$	&+dl+b&	\underline{2.61E-03}	$\pm$	1.10E-03	&	\underline{1.89E-03}	$\pm$	5.39E-04	&	\underline{1.72E-03}	$\pm$	4.14E-04	\\	
			&+dl--b&	\underline{2.73E-03}	$\pm$	1.07E-03	&	\underline{2.53E-03}	$\pm$	5.27E-04	&	\underline{3.94E-03}	$\pm$	4.25E-04	\\	
			&--dl+b&	\underline{3.51E-03}	$\pm$	1.05E-03	&	4.87E-03	$\pm$	6.33E-04	&	6.69E-03	$\pm$	3.76E-04	\\	
			&--dl--b&	3.78E-03	$\pm$	9.96E-04	&	4.96E-03	$\pm$	6.20E-04	&	6.73E-03	$\pm$	3.94E-04	\\	\hline
			
		\end{tabular}
		\label{tab2}
	\end{center}
\end{table}

From Tables 1--3 can be seen that for nonlinear functions all RVFL configurations (+dl+b, +dl--b, --dl+b and --dl--b) produce very similar results. Even in the case of NLF+L where TF contains a significant linear component. Only in four cases out of 81, the errors were slightly lower than for corresponding --dl--b configurations. These cases are: Gu +dl+b for NL, Gu --dl+b for NL, G$\alpha$ +dl+b for NLF+L, and G$\alpha$ +dl--b for NLF+L. Note that for 2-dimensional NL, --dl--b configurations gave lower errors than other configurations for each method of generating random parameters.

The optimal numbers of hidden nodes (averaged over 100 trials in each case) are shown in Table 5. Note that for NL there is no difference in the optimal number of nodes between RVFL configurations. Differences appear for multidimensional TFs with fluctuations, NLF and NLF+L, when random parameters are generated using Gs or Gu. In these cases, the configurations with direct links (+dl) need less hidden nodes than those without direct links (--dl). This is maybe because the hyperplane introduced by the direct links is useful for modeling the linear parts of the TFs (see the linear TF regions near the corner $\mathbf{x}=(0, 0, ..., 0)$ in the middle and right panels of Fig. \ref{figF}). We can see from Table 5 that for multidimensional TFs with fluctuations G$\alpha$ needs more nodes than Gs and Gu. But it was observed that also with a small number of nodes, G$\alpha$ still outperformed Gs and Gu in accuracy. Adding nodes led to decreasing in error for G$\alpha$, while for Gs and Gu increasing in error was observed at the same time \cite{Dud19}. This can be related to overfitting caused by the steeper nodes generated by Gs and Gu then by G$\alpha$, where the node slope angles are distributed uniformly. This phenomenon needs to be explored in detail on other TFs.
  
\begin{table}[]
	\setlength {\tabcolsep}{4pt}
	\caption{Optimal numbers of hidden nodes.}
	\begin{center}
		\begin{tabular}{|cc|ccc|ccc|ccc|ccc|}
			\hline
			\multicolumn{2}{|c|}{} &\multicolumn{3}{|c|}{NL} &	\multicolumn{3}{|c|}{NLF} &\multicolumn{3}{|c|}{NLF+L} &\multicolumn{3}{|c|}{L}\\	
			&$n$ &2	&	5	&10	&2	&	5	&10	&2	&5	&10	& 2	&5	&10	\\	\hline
			Gs	&+dl+b&	20	&	200	&	987	&	849	&	41	&	48	&	804	&	15	&	23	&	1.35	&	1.07	&	1.10	\\	
			&+dl--b&	20	&	200	&	988	&	849	&	41	&	51	&	809	&	17	&	25	&	1.62	&	1.04	&	1.19	\\	
			&--dl+b&	20	&	200	&	983	&	852	&	49	&	61	&	801	&	24	&	39	&	5.18	&	5.98	&	15.59	\\	
			&--dl--b&	20	&	200	&	982	&	852	&	50	&	62	&	800	&	25	&	40	&	6.62	&	7.07	&	20.80	\\	\hline
			Gu	&+dl+b&	19	&	232	&	963	&	638	&	44	&	53	&	476	&	17	&	26	&	1.14	&	1.08	&	1.17	\\	
			&+dl--b&	20	&	233	&	968	&	637	&	45	&	54	&	472	&	19	&	25	&	1.70	&	1.05	&	1.02	\\	
			&--dl+b&	20	&	228	&	968	&	641	&	51	&	58	&	483	&	26	&	34	&	4.12	&	5.58	&	13.50	\\	
			&--dl--b&	20	&	226	&	973	&	639	&	53	&	60	&	480	&	28	&	36	&	5.81	&	6.63	&	19.90	\\	\hline
			G$\alpha$	&+dl+b&	20	&	199	&	898	&	395	&	941	&	940	&	314	&	927	&	929	&	1.02	&	1.10	&	1.06	\\	
			&+dl--b&	20	&	200	&	901	&	395	&	941	&	940	&	315	&	930	&	930	&	1.63	&	5.21	&	25.87	\\	
			&--dl+b&	20	&	200	&	898	&	391	&	940	&	941	&	314	&	927	&	940	&	4.33	&	27.00	&	83.20	\\	
			&--dl--b&	20	&	200	&	901	&	391	&	940	&	941	&	314	&	929	&	940	&	5.69	&	30.00	&	82.90	\\	\hline

		\end{tabular}
		\label{tab2e}
	\end{center}
\end{table}

Table 4 shows the results for linear TF. This TF can be modeled with only direct links and bias. So, the hidden layer is unnecessary. Note that the optimal number of hidden nodes for the +dl+b configurations is around one (see Table 5) which is the minimum value of $m$ in our tests. The results for configurations without direct links (--dl) for L are usually much worse than those with direct links. Only for variants Gs and Gu at $ n=5 $ the errors were at a similar level for all network configurations. In the --dl configurations the linear TF is modeled with sigmoids and overfitting is a real threat when training data is noisy as in our case. Using only direct links and bias prevents overfitting for linear TFs. But it should be noted that for linear TFs we do not need to use NNs. Simple linear regression is a better choice. Moreover, linear TFs are rare in practice.                

Note that for highly nonlinear TFs such as NLF and NLF+L, G$\alpha$ ensures much more accurate fitting than other methods of generating random parameters (see Tables 2 and 3). For low dimensional TFs with fluctuations, Gu was more accurate than Gs. This is because, for low $n$, Gs generates many sigmoids that are saturated in the input hypercube and thus they are useless for modeling fluctuations. This phenomenon decreases with $n$ (see \cite{Dud19}).         
           
\section{Conclusion}

In this work, we investigate whether direct links and an output node bias are necessary in RVFL for regression problems. RVFL can be decomposed into a linear component represented by the direct links, a nonlinear component represented by the hidden nodes and a bias term. The experimental study showed that nonlinear target functions can be modeled with only nonlinear component. The fitting errors with and without direct links and bias in these cases were at a similar level. The linear component and bias term, if needed, can be replaced by hidden nodes. The direct links seem to be useful for modeling the target functions with linear regions. In our simulations modeling of such functions, NLF and NLF+L, required less hidden nodes when direct links were also used. This issue requires further research with target functions of different nature.  

In our study, we used three methods of generating random parameters of hidden nodes. The most sophisticated method proposed recently in the literature, G$\alpha$, was the most accurate especially for highly nonlinear target functions.


\begin{thebibliography}{8}
	
	\bibitem{Pri15}
	Principe, J.,Chen, B.: Universal approximation with convex  optimization: Gimmick or reality? IEEE Comput Intell Mag \textbf{10}, 68--77 (2015)
	
	\bibitem{Sch92}
	Schmidt, W.F., Kraaijveld, M.A., Duin, R.P.W.: Feedforward neural networks with random weights. In: Proc. 11th IAPR International Conference Pattern Recognition Methodology and Systems, vol. II, pp. 1--4 (1992)
	
	\bibitem{Wan17}
	Wang, D., Li, M.: Stochastic configuration networks: Fundamentals and algorithms. IEEE Trans on Cybernetics \textbf{47}(10), 3466--3479 (2017).
	
	\bibitem{Zha16}
	Le Zhang, Suganthan, P.N.: A comprehensive evaluation of random vector functional link networks. Information Sciences \textbf{367–368}, 1094--1105 (2016).
	
	\bibitem{Pao92}
	Pao, Y.H., Takefuji, Y.: Functional-link net computing: theory, system architecture, and functionalities. IEEE Comput \textbf{25}(5), 76--79 (1992) 
	
	\bibitem{Ige95}
	Igelnik, B., Pao, Y.H.: Stochastic choice of basis functions in adaptive function approximation and the functional-link net. IEEE Trans. Neural Netw. \textbf{6}(6), 1320--1329 (1995).
	
	\bibitem{Dud19b}
	Dudek, G.: Generating random weights and biases in feedforward neural networks with random hidden nodes. Information Sciences \textbf{481}, 33--56 (2019)
	
	\bibitem{Hus99}
	Husmeier, D.: Random vector functional link (RVFL) networks. In:  Neural Networks for Conditional Probability Estimation: Forecasting Beyond Point Predictions, chapter 6. Springer (1999).
	
	\bibitem{Zha16b}
	Le Zhang, Suganthan, P.N.: A survey of randomized algorithms for training neural networks. Information Sciences \textbf{364--365}, 146--155 (2016).
	
	\bibitem{Cao18}
	Cao, W., Wang, X., Ming, Z., Gao, J.: A review 	on neural networks with random weights. Neurocomputing \textbf{275}, 278--287 (2018).
	
	\bibitem{Pao94}
	Pao, Y.H., Park, G., Sobajic, D.: Learning and generalization characteristics of the random vector functional-link net. Neurocomputing \textbf{6}(2), 163--180 (1994).
		
	\bibitem{Dud19}
	Dudek, G.: Generating random parameters in feedforward neural networks with random hidden nodes: Drawbacks of the standard method and how to improve it. ArXiv:1908.05864 (2019).
	
	\bibitem{Tyu06}
	Tyukin, I., Prokhorov, D.: Feasibility of random basis function approximators for modeling and control. In: Proc. IEEE International Symposium on Intelligent Control, pp. 1391--1396 (2009).
	
	\bibitem{Dud19c}
	Dudek, G.: Improving randomized learning of feedforward neural networks by appropriate generation of random parameters. In: Advances in Computational Intelligence. 15th International Work-Conference on Artificial Neural Networks IWANN 2019. LNCS 11506, Springer, pp. 517--530 (2019). 
	
\end{thebibliography}
\end{document}